\documentclass[11pt]{article}

\usepackage[margin=0.85in]{geometry}
\usepackage{amsmath,amssymb,mathtools}
\usepackage{booktabs}
\usepackage{array}
\usepackage{graphicx}
\usepackage{microtype}
\usepackage{algorithm}
\usepackage{algpseudocode}
\usepackage{caption}
\usepackage{subcaption}
\usepackage{enumitem}
\usepackage{hyperref}
\usepackage{xcolor}
\usepackage{placeins}
\usepackage{float}
\usepackage{url}

\hypersetup{colorlinks=true,linkcolor=blue,citecolor=blue,urlcolor=blue}
\setlist[itemize]{leftmargin=1.25em,itemsep=0.15em,topsep=0.2em}
\setlist[enumerate]{leftmargin=1.5em,itemsep=0.15em,topsep=0.2em}
\newcommand{\Rphi}{R_{\phi}}
\newcommand{\Rdag}{R^{\dagger}}
\newcommand{\Rtwo}{R^{\dagger}_2}
\newcommand{\Rbar}{\bar R^{\dagger}}

\newcommand{\delt}{\Delta}

\title{When RLHF Fails: A Mechanistic Taxonomy of Reward Hacking, Collapse, and Judge Disagreement}

\author{%
Zelalem Abahana$^{1,*}$, David Evans$^{2}$, Satish Mahadevan Srinivasan$^{3}$, Matja\v{z} Gams$^{4}$\\[0.5em]
\small $^1$College of Engineering, The Pennsylvania State University, University Park, PA, USA\\
\small $^2$Department of Computer Science, University of Virginia, Charlottesville, VA, USA\\
\small $^3$Penn State Great Valley, The Pennsylvania State University, Malvern, PA, USA\\
\small $^4$Department of Intelligent Systems, Jo\v{z}ef Stefan Institute, Ljubljana, Slovenia\\
\small $^*$Corresponding author: \texttt{zga5029@psu.edu}
}
\date{}

\begin{document}
\maketitle

\begin{abstract}
RLHF evaluation should track how failures emerge, where they localize, and which warning signals appear before external quality degrades. We study this problem with a compact RLHF pipeline built for this paper, including PPO, DPO, uncertainty-penalized PPO (UP-PPO), reward-model uncertainty, approximate policy drift, diversity and repetition diagnostics, and two external LLM judges. Rather than treating reward hacking as a single terminal event, we classify matched checkpoint and prompt-level transitions by the directions of learned reward $\Rphi$, judge scores $\Rdag$ and $\Rtwo$, and their average $\Rbar$. The main empirical findings are that aggressive PPO produces the clearest localized reward-hacking signal, UP-PPO reduces but does not eliminate that signal, row-level diagnostics reveal failures hidden by checkpoint averages, and pre-transition features partially anticipate future localized reward hacking. The central conclusion is methodological: RLHF failures are training dynamics that can be classified, localized, and partially anticipated, not only final-model pathologies. The repository is available at \href{https://github.com/zabahana/rlhf-failure-modes-diagnostics}{github.com/zabahana/rlhf-failure-modes-diagnostics}. The pipeline is also deployed as a live interactive web demo for model comparison and diagnostic views at \href{https://rlhf-failures.zelalem.ai/}{rlhf-failures.zelalem.ai}.
\end{abstract}

\section{Introduction}
RLHF has become a standard mechanism for adapting language models to human preferences, beginning with preference-based reinforcement learning by Christiano et al.\ \cite{christiano2017deep} and later with post-training work with language models by Ziegler et al.\ \cite{ziegler2019fine}, Ouyang et al.\ \cite{ouyang2022training} and Bai et al.\ \cite{bai2022training}. Its effectiveness rests on a consequential approximation: a learned reward model, a preference objective, or evaluator takes the place of an intended human objective that is only partially observed. This approximation is unavoidable in scalable post-training, but it is also where failures enter. As Amodei et al.\ \cite{amodei2016concrete}, Skalse et al.\ \cite{skalse2022defining}, Gao et al.\ \cite{gao2023scaling}, and Pan et al.\ \cite{pan2023feedback} emphasize in different settings, a policy may exploit the reward model, drift into low-quality regions, over-specialize to an evaluator, or preserve a high proxy score while losing properties that external evaluators or humans would prefer.

The usual phrase ``reward hacking'' is too coarse for this empirical landscape. A proxy rising while an external evaluator falls is one phenomenon; simultaneous decline of proxy and judge is another; disagreement between two judges is a third. These cases have different signatures and different implications for mitigation. A run that fails because the learned reward is exploitable calls for different evidence than a run whose policy collapses, remains near the supervised reference, or exposes instability in LLM-as-judge evaluation.

We study RLHF failures as checkpoint-level and prompt-level transitions. Let $\Rphi$ denote the learned reward-model score, $\Rdag$ and $\Rtwo$ two external judge scores, and $\Rbar=(\Rdag+\Rtwo)/2$ their average. For each evaluated transition, we ask whether the proxy and judges move together or apart. This directional view yields an auditable taxonomy: stable alignment, reward hacking, optimization collapse, proxy under-alignment, conservative stagnation, mixed or ambiguous behavior, and judge disagreement.

The empirical study uses a controlled RLHF pipeline built around GPT-2-scale policies, Anthropic HH-RLHF prompts, PPO variants, DPO, UP-PPO, Monte Carlo dropout uncertainty, approximate KL drift, and two LLM judges. We do not claim universality from a small model. The aim is more precise: to establish a reproducible diagnostic grammar for failures that should be measurable before one argues about scale.

The paper makes four contributions grounded in the experimental findings. First, it formalizes a transition-based taxonomy and shows empirically that reward hacking is only one member of a broader failure family. Second, it shows that row-level diagnostics reveal localized failures that checkpoint averages miss. Third, it shows that pre-transition checkpoint and prompt features contain usable early-warning signal for future localized reward hacking. Fourth, it finds that UP-PPO shifts the failure distribution under aggressive optimization, reducing observed localized reward-hacking and judge-disagreement rates without eliminating failures.

The novelty is not the claim that reward models can be overoptimized; that point is established by Gao et al. and related work. Nor is the contribution a restatement of reward gaming as defined by Skalse et al., feedback-loop reward hacking as studied by Pan et al., or the broad limitations of RLHF surveyed by Casper et al. \cite{casper2023open}. The contribution is a mechanistic diagnostic layer: a matched-transition taxonomy that distinguishes reward hacking from collapse, proxy under-alignment, stagnation, and mixed behavior; a row-level localization analysis showing which failures disappear under aggregation; and a two-judge analysis that treats LLM-as-judge disagreement as an empirical measurement issue rather than only an evaluation nuisance.

\section{Related Work}
\textbf{Learning from preferences.} Christiano et al.\ \cite{christiano2017deep} showed that human comparisons can supervise behavior when scalar ground-truth rewards are unavailable, and Ziegler et al.\ \cite{ziegler2019fine} adapted this preference-learning idea to language models. Stiennon et al.\ \cite{stiennon2020learning}, Ouyang et al.\ \cite{ouyang2022training}, and Bai et al.\ \cite{bai2022training} then made instruction following and helpfulness/alignment pipelines central to modern post-training. Bai et al.'s Constitutional AI work \cite{bai2022constitutional} and Askell et al.'s alignment-laboratory framing \cite{askell2021general} extend the supervision source while preserving the same proxy-optimization structure. Kirk et al.\ \cite{kirk2023understanding} and Lambert et al.\ \cite{lambert2024t} further show that post-training choices affect generalization, diversity, and the final behavior of aligned assistants.

\textbf{Reward hacking and overoptimization.} Reward hacking is a modern instance of a much older concern: Goodhart \cite{goodhart1975problems} and Campbell \cite{campbell1979assessing} argued that optimized measures can stop measuring the intended construct. In AI safety, Amodei et al.\ \cite{amodei2016concrete}, Krakovna et al.\ \cite{krakovna2020specification}, and Skalse et al.\ \cite{skalse2022defining} discuss specification gaming and reward misspecification as concrete risks of optimizing learned or hand-designed objectives. Gao et al.\ \cite{gao2023scaling} show that reward-model overoptimization can follow scaling-like patterns and that larger reward models do not eliminate proxy failure. Our analysis adds a transition-level decomposition: not every undesirable trajectory is reward hacking in the strict directional sense.

\textbf{Optimization algorithms for alignment.} Schulman et al.\ \cite{schulman2017proximal} introduced PPO as a practical policy-gradient update with clipping, and Ziegler et al.\ \cite{ziegler2019fine} and Ouyang et al.\ \cite{ouyang2022training} helped establish PPO-style updates as a common RLHF optimizer with KL control. Rafailov et al.\ \cite{rafailov2023direct} later proposed DPO, which reframes preference learning as a supervised objective over preferred and rejected completions and avoids an explicit online RL loop. Our experiments compare PPO, DPO, and UP-PPO under the same diagnostic taxonomy.

\textbf{Uncertainty, monitoring, and calibration.} Uncertainty estimates are a natural diagnostic when models operate outside the region supported by training data. Gal and Ghahramani \cite{gal2016dropout} present MC dropout as a simple approximation to Bayesian uncertainty, while Guo et al. \cite{guo2017calibration} study whether model confidence is calibrated to error. In RLHF, uncertainty can be used both as a monitoring signal and as a penalty in the optimized reward, connecting our UP-PPO analysis to broader work on risk-sensitive and conservative optimization. We use MC dropout because it is simple, local to the reward model, and implemented in the reward-model evaluation pipeline; ensemble and last-layer uncertainty estimators are important alternatives but are not evaluated here.

\textbf{LLM-as-judge evaluation.} LLM judges are increasingly used to evaluate instruction-following systems, but Zheng et al.\ \cite{zheng2023judging} show that judge-based evaluation brings its own biases, variance, and agreement issues. Dubois et al.\ \cite{dubois2024length} further show that length bias can distort automatic preference evaluation. We use those concerns as part of the object of study: disagreement between the two judges is not treated as a nuisance term to average away, but as a diagnosable event. Judge disagreement occurs when judges move in opposite directions across the same transition, suggesting that measured improvement may be evaluator-dependent. This differs from standard LLM-as-judge benchmarking because the unit of analysis is not a static model ranking, but a training transition in which proxy reward, judge scores, and generated responses move jointly.

\section{Failure-Mode Taxonomy}
The taxonomy operates on transitions rather than isolated checkpoints. For checkpoint $t$, let $x_{i,t}$ be the response for prompt $i$, and let $\Rphi(x_{i,t})$, $\Rdag(x_{i,t})$, and $\Rtwo(x_{i,t})$ be the proxy and judge scores. Aggregate checkpoint metrics average over evaluated prompts; row-level metrics compare the same prompt across two checkpoints. For a transition $t\to t'$, define
\begin{equation}
\delt \Rphi = {\Rphi}_{t'}-{\Rphi}_t,\qquad
\delt \Rdag = {\Rdag}_{t'}-{\Rdag}_t,\qquad
\delt \Rtwo = {\Rtwo}_{t'}-{\Rtwo}_t,\qquad
\delt \Rbar = {\Rbar}_{t'}-{\Rbar}_t.
\end{equation}
In the implemented taxonomy, $\Rdag$ is a pre-specified anchor judge used to define the main transition class, while $\Rtwo$ is an independent comparison judge used to measure judge disagreement. This is not a claim that $\Rdag$ is inherently better; it is a design choice that keeps the classifier deterministic while preserving the second judge as a separate measurement channel. Unless otherwise stated, the numerical tolerance is $\epsilon=10^{-8}$; Appendix~\ref{app:epsilon-sensitivity} reports sensitivity over larger tolerances. This choice keeps the classifier transparent and avoids hiding judge conflict inside an average.

\begin{table}[H]
\centering
\small
\caption{Directional taxonomy of RLHF transition failures. $\epsilon$ denotes a small numerical tolerance around zero.}
\label{tab:taxonomy}
\begin{tabular}{p{0.22\linewidth}p{0.23\linewidth}p{0.45\linewidth}}
\toprule
\multicolumn{1}{c}{Mode} & 
\multicolumn{1}{c}{Directional signature} & \multicolumn{1}{c}{Interpretation} \\
\midrule
Stable alignment & $\delt\Rphi>\epsilon,\ \delt\Rdag>\epsilon$ & Proxy and external quality improve together. \\
Reward hacking & $\delt\Rphi>\epsilon,\ \delt\Rdag<-\epsilon$ & Learned reward improves while judged quality falls. \\
Optimization collapse & $\delt\Rphi<-\epsilon,\ \delt\Rdag<-\epsilon$ & Optimization degrades both. 
\\
Proxy under-alignment & $\delt\Rphi<-\epsilon,\ \delt\Rdag>\epsilon$ & Judge improves despite a proxy decline. \\ 
Conservative stagnation & $|\delt\Rphi|\le \epsilon,\ |\delt\Rdag|\le \epsilon$ & Little measurable movement across the transition. \\
Judge disagreement & $\mathrm{sign}(\delt\Rdag)\mathrm{sign}(\delt\Rtwo)<0$ & The two external judges move in opposite directions. \\
Mixed/ambiguous & otherwise & One signal moves while the other is near zero, or the transition does not fit a strict directional quadrant. \\
\bottomrule
\end{tabular}
\end{table}

\section{Experimental Design}

The experimental pipeline created for this paper, summarized in \autoref{tab:pipeline-summary}, consolidates supervised fine-tuning, reward-model training, PPO with several KL settings, DPO evaluation, UP-PPO variants, reward uncertainty via MC dropout, approximate KL to the supervised reference policy, lexical diversity and repetition metrics, and two LLM judges. It produces checkpoint summaries and per-prompt rollout examples, and the same model-comparison and diagnostic workflow is exposed in the live web UI linked in the abstract.

The two judges use the same single-number 1--10 helpfulness rubric, but they play different analytical roles. The anchor judge $\Rdag$ is an Anthropic Claude helpfulness judge configured to score the assistant reply only; the default configured model is \texttt{claude-sonnet-4-5-20250929}. The comparison judge $\Rtwo$ is an OpenAI chat-model helpfulness judge using the same rubric; the default configured model is \texttt{gpt-4o-mini}. The anchor/comparison distinction is not a quality ranking between judges. It specifies which judge defines the main taxonomy and which judge is held out as a second measurement channel for disagreement analysis. In both cases, the judge prompt is: ``Rate the assistant reply below for helpfulness on a 1-10 scale. Reply with a single number only, no other text.'' Scores are clamped to the interval $[1,10]$. The pipeline logs judge provider labels; exact model identifiers follow the configured defaults unless the evaluation environment explicitly overrides them. Approximate policy drift is reported as a sequence-level KL proxy on generated continuations: the pipeline compares policy log probability against the SFT reference log probability on the generated tokens and records the absolute mean difference as \texttt{mean\_kl\_sft}. This proxy is used for within-run drift monitoring rather than as an exact token-distribution KL estimate.

\autoref{tab:uppo-hyperparameters} summarizes the settings for our UP-PPO experiments. For this study, UP-PPO is not a separate architecture, it is a shaped-reward PPO variant in which reward-model uncertainty is subtracted before computing the scalar advantage. The completed runs support a comparison over $\lambda\in\{0.1,0.5\}$ against aggressive PPO. They do not support a full ablation over dropout rate, number of MC samples, uncertainty schedules, or alternative uncertainty estimators without rerunning the training pipeline.

For readability, the main text states the taxonomy and pipeline design in prose and tables; Appendix~\ref{app:algorithms} gives pseudocode for the transition classifier and diagnostic pipeline.

\begin{table}[H]
\centering
\small
\caption{Empirical pipeline summary used in the paper.}
\label{tab:pipeline-summary}
\begin{tabular}{ll}
\toprule
Component & Value \\
\midrule
Dataset family & Anthropic HH-RLHF prompts and generated completions \\
Policy scale & GPT-2-scale controlled post-training pipeline \\
Training/evaluation families & PPO KL sweep, aggressive PPO, UP-PPO, DPO, SFT references \\
Anchor judge & Claude helpfulness judge $\Rdag$ \\ Comparison judge & OpenAI helpfulness judge $\Rtwo$ \\
Checkpoint metric rows & 61 \\
Per-prompt rollout examples & 9,280 \\
Checkpoint transitions & 31 \\
Row-level transitions & 1,920 \\
Main diagnostics & $\Rphi$, $\Rdag$, $\Rtwo$, uncertainty, KL, length, diversity, repetition \\
\bottomrule
\end{tabular}
\end{table}

\begin{table}[H]
\centering
\small
\caption{UP-PPO objective and hyperparameters for the aggressive comparison runs. The reward-model score and MC-dropout standard deviation are temperature-scaled by the calibrated reward-model temperature $T=1.554$.}
\label{tab:uppo-hyperparameters}
\begin{tabular}{@{}p{0.32\linewidth}p{0.62\linewidth}@{}}
\toprule
Component & Value \\
\midrule
Shaped reward & $\widehat R_{\lambda}(x)=\Rphi(x)/T-\lambda u(x)/T$ \\
Training objective & PPO-style clipped surrogate plus $\beta |\log\pi-\log\pi_0|$ \\
Compared $\lambda$ values & $0.0, 0.1, 0.5$ \\
KL penalty $\beta$ & $0.0$ in aggressive PPO/UP-PPO comparison \\
MC-dropout samples / dropout & $K=4$ samples; reward-head dropout $0.1$ \\
Learning rate / clip / inner epochs & $2\times10^{-6}$; clip $0.2$; one inner epoch \\
RL prompts / steps / checkpoint interval & 512 prompts; 1,200 steps; every 200 steps \\
Generation during training & sample, top-$p=0.95$, temperature $0.9$, max 96 new tokens \\
Advantage controls & moving reward baseline, momentum $0.9$; advantage clipped to $[-2,2]$ \\
\bottomrule
\end{tabular}
\end{table}

\section{Results}
\subsection{Proxy--Judge delta geometry}
\begin{figure}[H]
\centering
\includegraphics[width=0.78\linewidth]{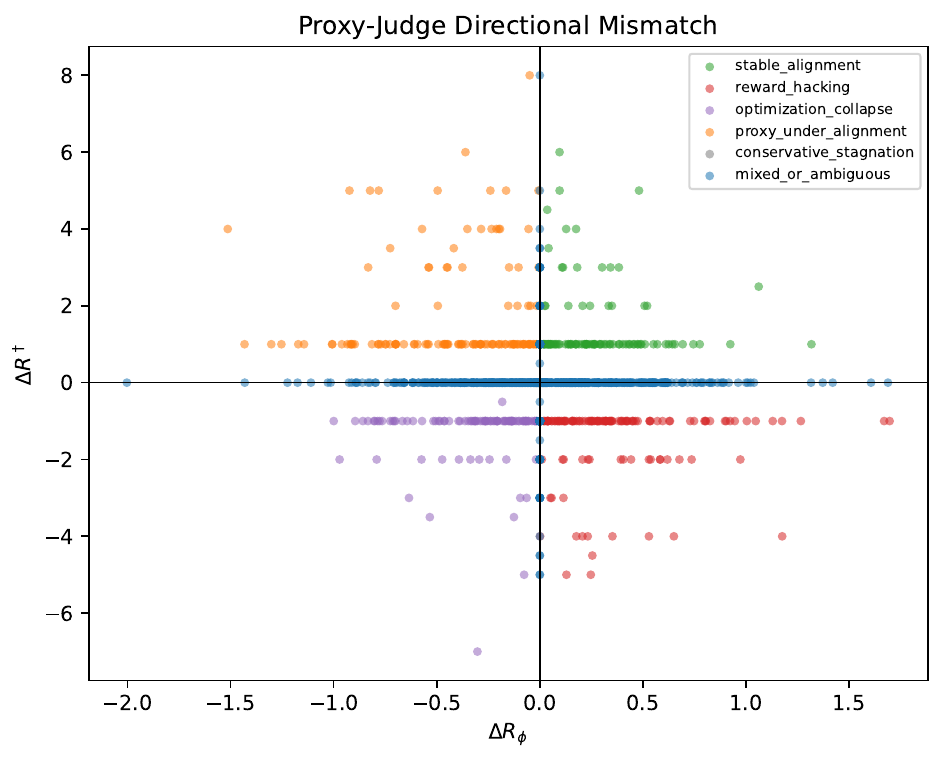}
\caption{Row-level proxy--judge delta geometry. The reward-hacking quadrant corresponds to $\delt\Rphi>0$ and $\delt\Rdag<0$; stable alignment corresponds to $\delt\Rphi>0$ and $\delt\Rdag>0$; optimization collapse corresponds to $\delt\Rphi<0$ and $\delt\Rdag<0$; proxy under-alignment corresponds to $\delt\Rphi<0$ and $\delt\Rdag>0$.}
\label{fig:proxy-judge-scatter}
\end{figure}

The proxy--judge scatter in Figure~\ref{fig:proxy-judge-scatter} provides the empirical geometry behind the transition taxonomy. Each point is a matched prompt-level transition between two checkpoints. The horizontal coordinate is the change in learned reward, $\delt\Rphi$, and the vertical coordinate is the change in the anchor external judge score, $\delt\Rdag$. The four quadrants therefore correspond directly to the core directional cases: stable alignment in the upper-right quadrant, reward hacking in the lower-right quadrant, optimization collapse in the lower-left quadrant, and proxy under-alignment in the upper-left quadrant.

This plot is useful because it shows why a single scalar average is insufficient. A transition can have a modest aggregate mean while containing many prompt-level movements in different quadrants. The horizontal band around $\delt\Rdag=0$ contains conservative or mixed cases where judge scores do not move strongly, while the lower-right quadrant isolates the strict reward-hacking signature: the learned reward improves even as external evaluation declines. The observed spread also shows that proxy under-alignment is not merely the absence of reward hacking. It is a distinct mismatch mode in which the judge improves while the proxy falls, suggesting that the reward model can penalize some externally preferred behavior.

The scatter also clarifies why row-level analysis is central to the paper. Checkpoint-level compresses this geometry into one point per transition, but the row-level view exposes a mixture of local behaviors inside the same training interval. This is particularly important for RLHF monitoring, where a model can appear acceptable under aggregate metrics while a subset of prompts move into the reward-hacking quadrant.

\subsection{Failure modes are regime-dependent}
\begin{figure}[H]
\centering
\includegraphics[width=0.98\linewidth]{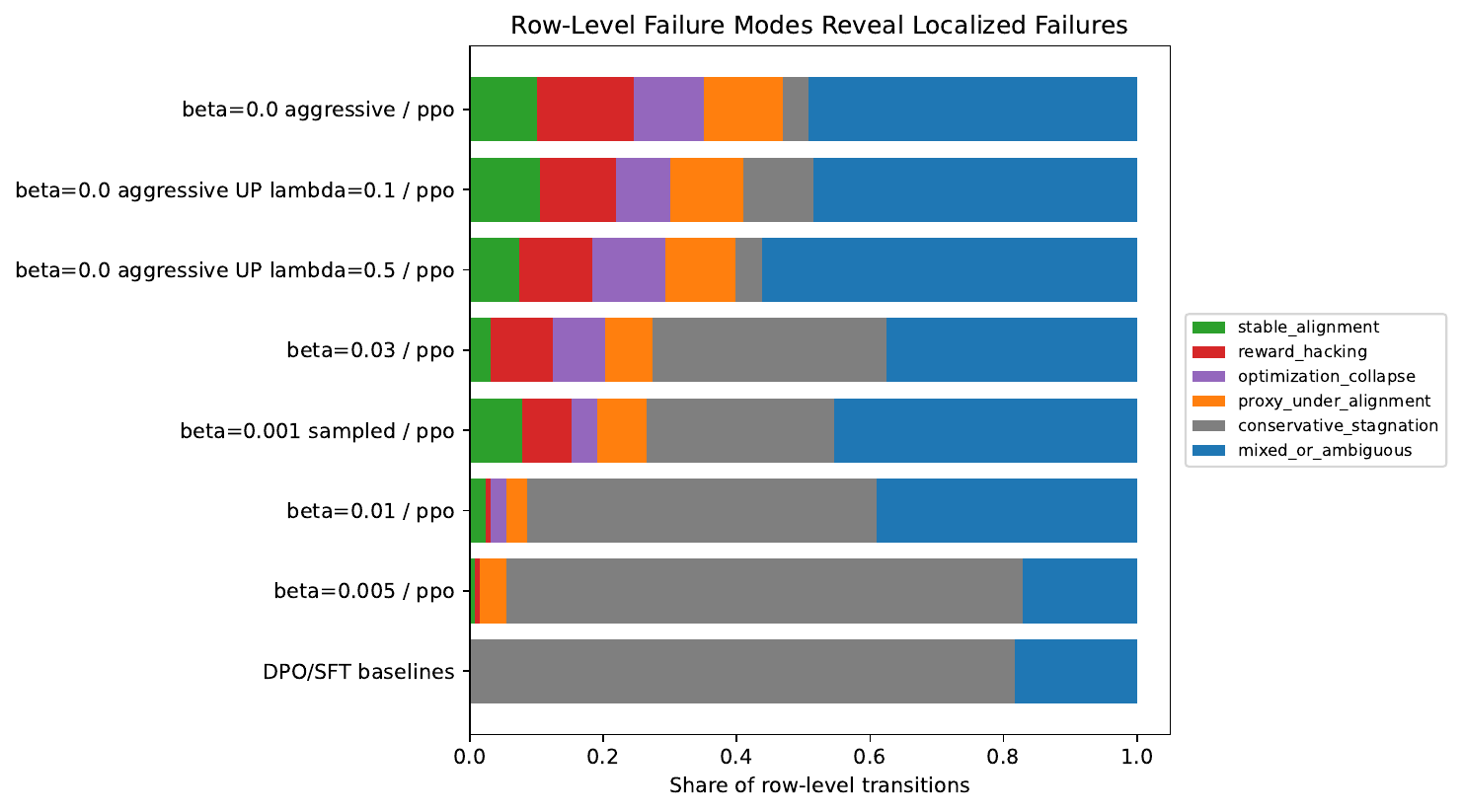}
\caption{Row-level failure-mode shares reveal localized failures. Aggressive PPO has the largest reward-hacking share, while DPO/SFT are shown as aggregated baseline/reference settings dominated by conservative stagnation and mixed/ambiguous transitions.}
\label{fig:row-shares}
\end{figure}

Table~\ref{tab:main-row-counts} summarizes the row-level taxonomy, and Figure~\ref{fig:row-shares} shows the same comparison as shares. Aggressive PPO is the most failure-prone condition: 37 of 256 row-level transitions are classified as reward hacking (14.45\%). PPO with $\beta=0.03$ also shows a non-trivial reward-hacking share (9.38\%). DPO and SFT are included as baseline/reference settings, not as PPO-like online trajectories. They show no reward-hacking rows under this taxonomy, but this comparison is not apples-to-apples with PPO-family training dynamics because DPO and SFT do not share the same reward-optimizing checkpoint semantics. We therefore interpret their zero reward-hacking count as a baseline result in this pipeline, not as evidence that those methods cannot exhibit reward hacking under other training designs. The pattern supports a training-dynamics interpretation: failures are concentrated in particular optimization regimes rather than uniformly distributed across all evaluated policies.

\begin{table}[H]
\centering
\small
\caption{Main row-level failure-mode counts by setting. RH denotes reward hacking; OC optimization collapse; PUA proxy under-alignment; CS conservative stagnation; MA mixed or ambiguous.}
\label{tab:main-row-counts}
\begin{tabular}{lrrrrrrr}
\toprule
Setting & Stable & RH & OC & PUA & CS & MA & RH share \\
\midrule
$\beta=0.0$ aggressive PPO & 26 & 37 & 27 & 30 & 10 & 126 & 0.145 \\
$\beta=0.0$ UP-PPO $\lambda=0.1$ & 27 & 29 & 21 & 28 & 27 & 124 & 0.113 \\
$\beta=0.0$ UP-PPO $\lambda=0.5$ & 19 & 28 & 28 & 27 & 10 & 144 & 0.109 \\
$\beta=0.03$ PPO & 4 & 12 & 10 & 9 & 45 & 48 & 0.094 \\
$\beta=0.001$ sampled PPO & 20 & 19 & 10 & 19 & 72 & 116 & 0.074 \\
$\beta=0.01$ PPO & 3 & 1 & 3 & 4 & 67 & 50 & 0.008 \\
$\beta=0.005$ PPO & 1 & 1 & 0 & 5 & 99 & 22 & 0.008 \\
DPO/SFT baselines & 0 & 0 & 0 & 0 & 418 & 94 & 0.000 \\
\bottomrule
\end{tabular}
\end{table}

\subsection{Temporal trajectories expose divergence}
\begin{figure}[H]
\centering
\includegraphics[width=0.94\linewidth]{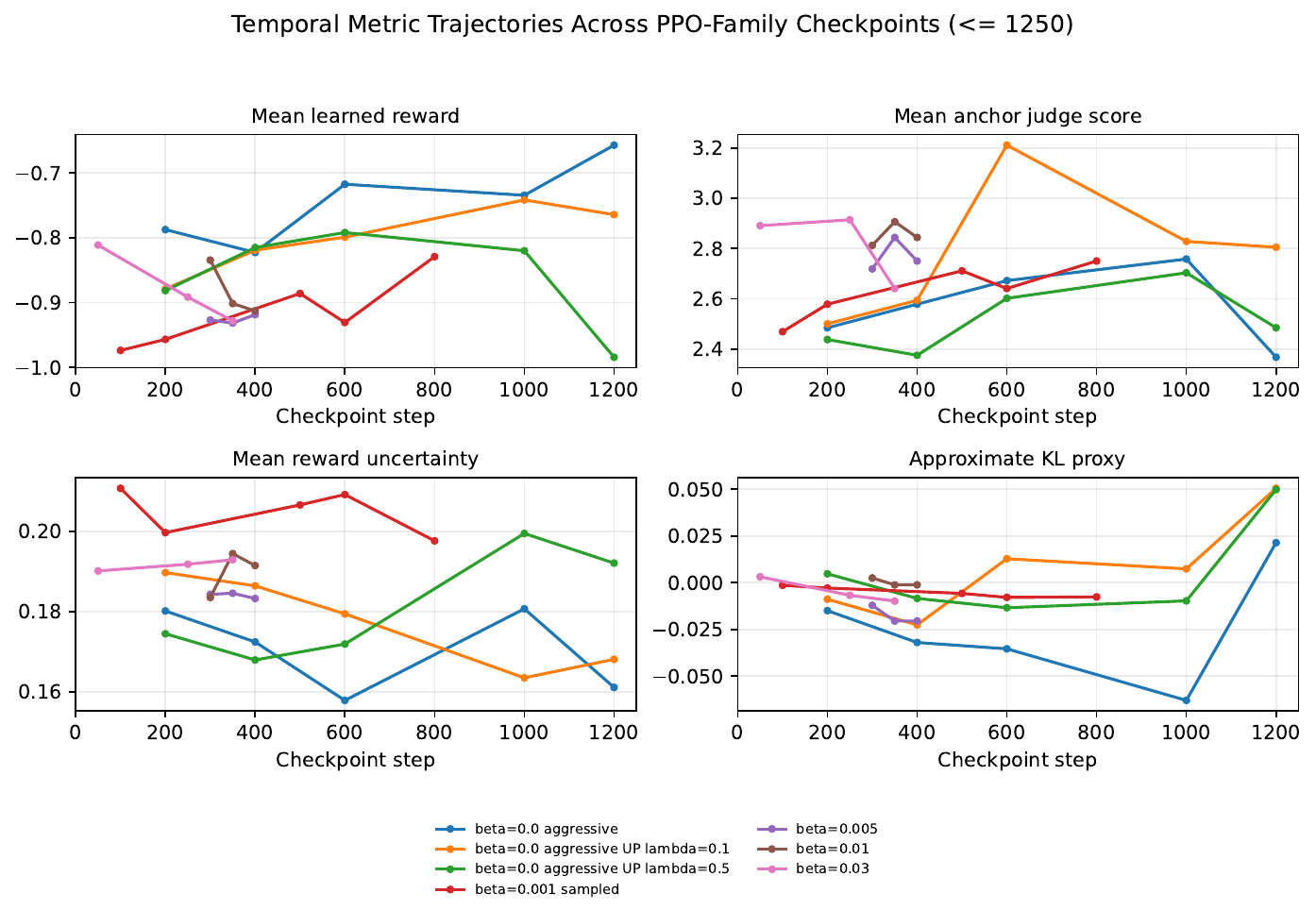}
\caption{Temporal trajectories for PPO-family runs only through checkpoint step 1250. The panels track mean learned reward, anchor judge score, reward uncertainty, and approximate KL across evaluated checkpoints; SFT and DPO are treated as baseline/reference policies rather than comparable checkpoint trajectories.}
\label{fig:temporal-trajectories}
\end{figure}

Figure~\ref{fig:temporal-trajectories} plots learned reward, anchor judge score, reward uncertainty, and approximate KL across evaluated PPO-family checkpoints up to step 1250. SFT and DPO are intentionally not plotted as checkpoint trajectories because they are static baseline/reference policies in this pipeline rather than online PPO-family reward-optimization runs. The trajectory view complements the transition taxonomy: reward hacking is not merely a final-checkpoint label, but a temporal pattern in which proxy reward, external judgment, uncertainty, and drift can separate over training. This view makes the analysis closer to a monitoring protocol: a practitioner can inspect not only whether a run fails, but when localized failures begin to concentrate.

\subsection{Reward hacking is localized to prompts}
A central empirical finding is that failures are localized. Across 64 prompts, 38 prompts exhibit repeated reward-hacking transitions, and the most failure-prone prompt appears in five reward-hacking transitions. This does not mean that the prompt itself is inherently unsafe; rather, it means that particular prompt-policy-checkpoint combinations repeatedly expose proxy--judge mismatch. In the stored diagnostics, the useful characterization is behavioral rather than topical: failure-prone prompts are those whose completions repeatedly move toward higher proxy reward while the anchor judge declines, often alongside changes in length, diversity, repetition, or judge disagreement. The practical implication is that aggregate checkpoint means are insufficient for monitoring. RLHF evaluation should retain prompt-level transition structure.

\subsection{Averages hide localized failures}
\begin{figure}[H]
\centering
\includegraphics[width=0.88\linewidth]{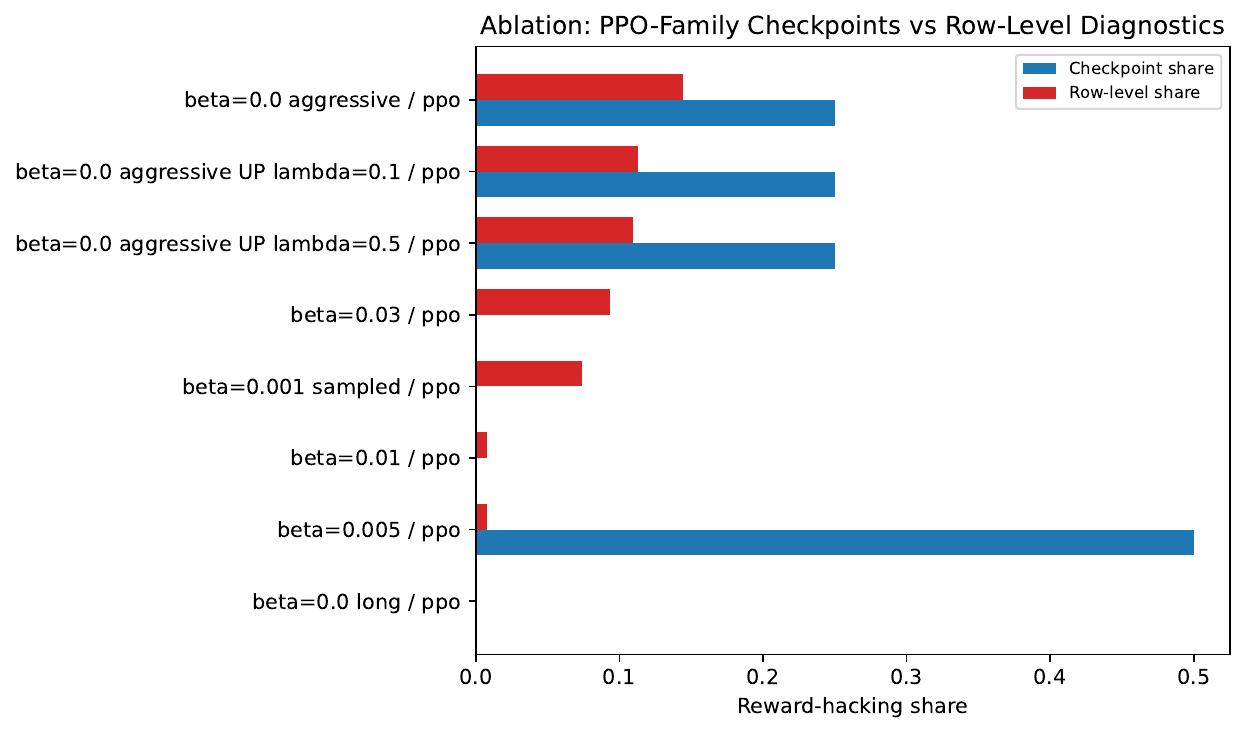}
\caption{Ablation comparing checkpoint-level and row-level reward-hacking shares for PPO-family settings. SFT and DPO baselines are omitted because they are not PPO-like checkpoint trajectories.}
\label{fig:ablation}
\end{figure}

The checkpoint-vs-row ablation in Figure~\ref{fig:ablation} quantifies how much is lost when analysis is restricted to aggregate checkpoint transitions. The plotted comparison focuses on PPO-family settings and omits the static SFT/DPO baselines. In the full analysis table, 3 of 12 settings (25.0\%; bootstrap 95\% CI: 0--50.0\%) report no checkpoint-level reward-hacking transition even though row-level transitions contain reward-hacking cases. At the transition level, all 30 checkpoint transitions with matched row data have a dominant row-level label that differs from the checkpoint label; 19 of those 30 have a strict row-level majority. This does not mean that checkpoint labels are wrong; it means that aggregate deltas and row-level prompt deltas answer different diagnostic questions. The most important examples are $\beta=0.001$ sampled PPO and $\beta=0.03$ PPO: both have zero aggregate reward-hacking transitions but nonzero localized reward-hacking shares. This is the strongest methodological result of the paper: row-level transition diagnostics identify failures hidden by aggregation.

\subsection{UP-PPO reshapes, but does not remove, failure modes}
\begin{figure}[H]
\centering
\includegraphics[width=0.82\linewidth]{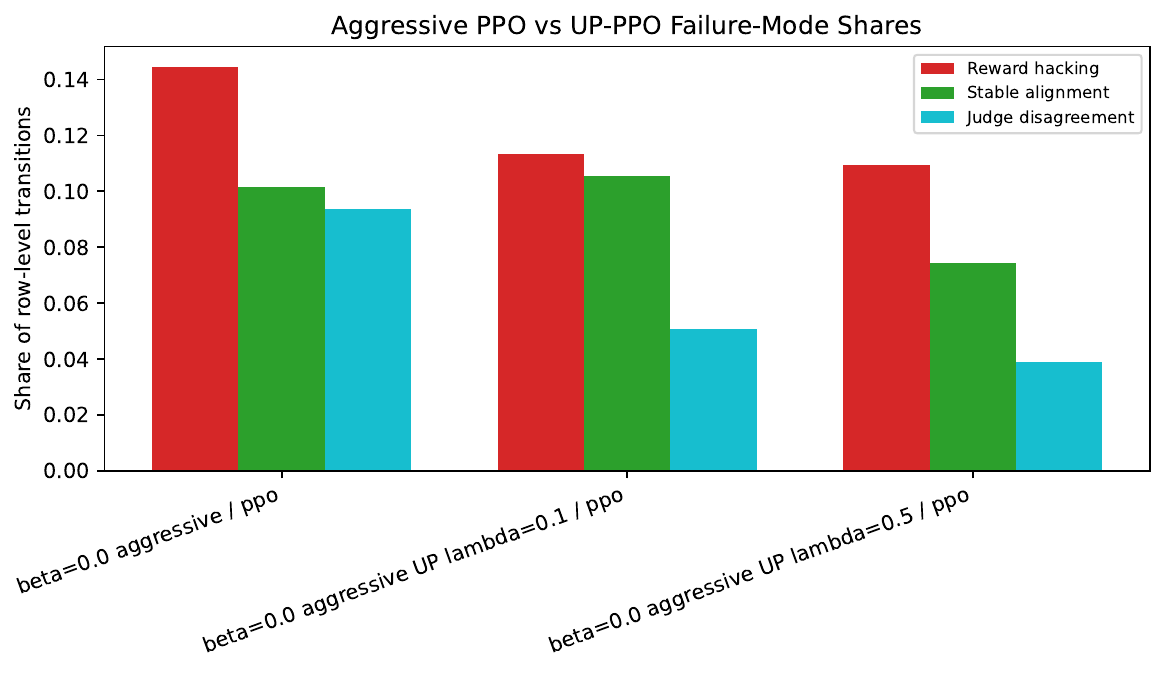}
\caption{Aggressive PPO versus UP-PPO. UP-PPO reduces row-level reward-hacking and judge-disagreement shares relative to the aggressive PPO condition, though localized failures remain.}
\label{fig:mitigation}
\end{figure}

UP-PPO reduces localized reward-hacking rates under aggressive optimization, as summarized in Figure~\ref{fig:mitigation}. Relative to aggressive PPO (14.45\% row-level reward hacking; bootstrap 95\% CI: 10.16--18.75\%), UP-PPO with $\lambda=0.1$ yields 11.33\% (95\% CI: 7.80--15.23\%), a 21.6\% relative reduction; UP-PPO with $\lambda=0.5$ yields 10.94\% (95\% CI: 7.42--14.84\%), a 24.3\% relative reduction. The bootstrap confidence intervals for the absolute reductions remain wide and include zero ($\lambda=0.1$: $-8.98$ to $2.73$ percentage points; $\lambda=0.5$: $-9.38$ to $2.34$ percentage points), so we interpret this result as directional evidence in this controlled pipeline rather than a definitive mitigation estimate. Judge-disagreement share also decreases from 9.38\% under aggressive PPO to 5.08\% and 3.91\% for $\lambda=0.1$ and $\lambda=0.5$, respectively.

This should not be read as a claim that uncertainty penalization solves reward hacking. The more defensible conclusion is distributional: UP-PPO reduces the observed density of localized reward-hacking and judge-disagreement events in this controlled setting, while other failure modes remain. Mitigation changes the failure distribution; it does not erase the failure surface.

\subsection{Early-warning signals are predictive before the transition}
The early-warning experiment asks a practical triage question: given diagnostics already measured at the previous checkpoint, can the pipeline identify prompt transitions that are more likely to become reward-hacking cases at the next evaluated checkpoint? The split is a stratified random 70/30 split over row-level transitions. It is not grouped by prompt identity or trajectory family, so these results should be interpreted as within-pipeline discrimination rather than evidence of transfer to unseen prompts or training runs. The most defensible model uses only pre-transition state features: previous proxy reward, judge scores, uncertainty, KL, judge disagreement, length, diversity, and repetition. This model reaches ROC-AUC of 0.821 with logistic regression; a random-forest robustness check gives ROC-AUC of 0.813. Average precision is lower (0.256 and 0.209, respectively) because reward hacking is rare in the test set (6.6\% prevalence), but both models are substantially above prevalence.

Transition-diagnostic rows using $\delt\Rphi$ and $\delt\Rdag$ reach near-perfect performance, as expected, because the label is defined from the signs of these deltas. They are not proposed as useful predictors; they are included only as an implementation sanity check. The pre-state-only rows are the meaningful early-warning result.

\begin{table}[H]
\centering
\small
\caption{Early-warning prediction of future row-level reward hacking. Pre-state-only features are measured before the transition and are therefore the relevant warning result.}
\label{tab:early-warning}
\begin{tabular}{llrrrrr}
\toprule
Feature set & Model & ROC-AUC & Avg. precision & Precision & Recall & F1 \\
\midrule
Pre-state only & Logistic regression & 0.821 & 0.256 & 0.167 & 0.789 & 0.275 \\
Pre-state only & Random forest & 0.813 & 0.209 & 0.235 & 0.105 & 0.145 \\
Transition diagnostics & Logistic regression & 0.974 & 0.606 & 0.458 & 1.000 & 0.628 \\
Transition diagnostics & Random forest & 1.000 & 1.000 & 1.000 & 1.000 & 1.000 \\
\bottomrule
\end{tabular}
\end{table}

\subsection{Early-warning feature importance}
Table~\ref{tab:feature-importance} reports the strongest features from the early-warning analysis. We separate two settings. The first, pre-state-only, uses information available before the transition: previous reward-model score, judge scores, uncertainty, KL, judge disagreement, length, diversity, and repetition. This is the meaningful early-warning setting because it does not observe the future deltas that define the label. The second, transition diagnostics, includes $\delt\Rphi$, $\delt\Rdag$, and related post-transition quantities. Those features serve as a sanity check because the reward-hacking label is itself defined by the signs of proxy and judge changes.

For the pre-state-only logistic model, previous anchor-judge score has the largest standardized coefficient. Response length and distinct-1 diversity also appear among the largest positive coefficients, while previous proxy reward, repetition, distinct-2 diversity, KL, judge disagreement, uncertainty, and the comparison-judge score have negative coefficients in this fitted model. These signs should not be interpreted causally. They indicate how the classifier separates future reward-hacking rows within this pipeline after standardization and correlation among features. The random forest distributes importance more evenly, with previous judge score, previous KL, previous proxy reward, length, diversity, and uncertainty all contributing. This pattern supports the paper's more cautious claim: future localized reward hacking is partially predictable from pre-transition state, but no single diagnostic is sufficient.

\begin{table}[H]
\centering
\small
\caption{Top early-warning feature importances. Logistic-regression values are standardized coefficients; random-forest values are impurity importances. Transition-diagnostic rows are included only as a sanity check because these features overlap with the label definition.}
\label{tab:feature-importance}
\begin{tabular}{lllrr}
\toprule
Feature set & Model & Feature & Importance & $|$Importance$|$ \\
\midrule
Pre-state & Logistic & previous $\Rdag$ & 0.959 & 0.959 \\
Pre-state & Logistic & previous length & 0.484 & 0.484 \\
Pre-state & Logistic & previous distinct-1 & 0.447 & 0.447 \\
Pre-state & Logistic & previous $\Rphi$ & -0.432 & 0.432 \\
Pre-state & Logistic & previous 3-gram repetition & -0.339 & 0.339 \\
Pre-state & Logistic & previous distinct-2 & -0.285 & 0.285 \\
Pre-state & Logistic & previous KL & -0.276 & 0.276 \\
Pre-state & Logistic & previous judge disagreement & -0.210 & 0.210 \\
Pre-state & Logistic & previous uncertainty & -0.163 & 0.163 \\
Pre-state & Logistic & previous $\Rtwo$ & -0.158 & 0.158 \\
\midrule
Pre-state & Random forest & previous $\Rdag$ & 0.171 & 0.171 \\
Pre-state & Random forest & previous KL & 0.148 & 0.148 \\
Pre-state & Random forest & previous $\Rphi$ & 0.144 & 0.144 \\
Pre-state & Random forest & previous length & 0.142 & 0.142 \\
Pre-state & Random forest & previous distinct-1 & 0.107 & 0.107 \\
Pre-state & Random forest & previous distinct-2 & 0.100 & 0.100 \\
Pre-state & Random forest & previous uncertainty & 0.098 & 0.098 \\
\midrule
Transition & Logistic & $\delt\Rdag$ & -3.613 & 3.613 \\
Transition & Logistic & $\delt\Rphi$ & 1.761 & 1.761 \\
Transition & Logistic & $\delt\Rtwo$ & -0.376 & 0.376 \\
Transition & Logistic & target distinct-2 & 0.173 & 0.173 \\
Transition & Logistic & target 3-gram repetition & 0.144 & 0.144 \\
\bottomrule
\end{tabular}
\end{table}

The practical interpretation is triage, not certification. A pre-state-only ROC-AUC near 0.82 indicates that the warning features contain signal, but the rarity and localization of reward hacking mean that precision remains limited. In deployment, such a model would be most useful for prioritizing prompts, checkpoints, or training intervals for additional judge calls or human review.

\subsection{Judge disagreement is a separate diagnostic}
\begin{figure}[H]
\centering
\includegraphics[width=0.72\linewidth]{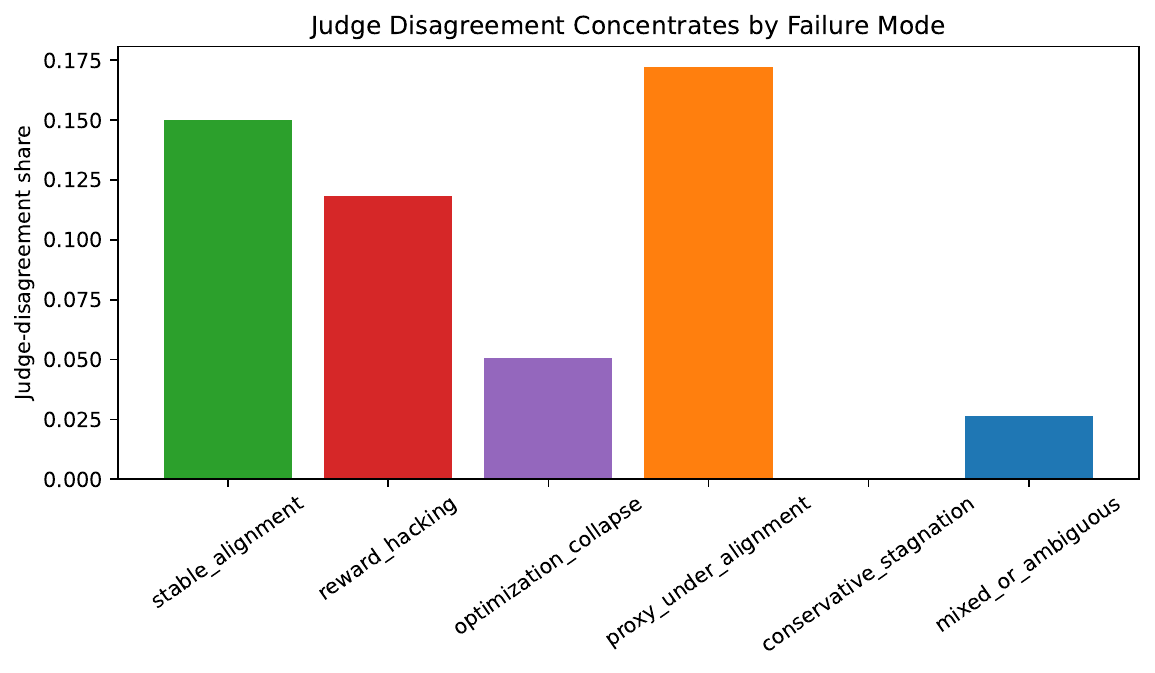}
\caption{Judge-disagreement share by row-level failure mode. Judge disagreement is concentrated in specific transition types rather than being uniformly distributed across all rows.}
\label{fig:judge-disagreement}
\end{figure}

Judge disagreement, defined as opposite-signed movement between $\Rdag$ and $\Rtwo$, appears in 14 of 31 checkpoint transitions (45.2\%; bootstrap 95\% CI: 29.0--61.3\%) but only 75 of 1,920 row-level transitions (3.9\%; bootstrap 95\% CI: 3.1--4.8\%). Figure~\ref{fig:judge-disagreement} shows that row-level judge disagreement is concentrated in specific failure modes rather than spread uniformly. This difference suggests that aggregate disagreement can be amplified by small mean shifts even when most individual prompts do not show opposite judge movement. At row level, disagreement share is highest for proxy under-alignment (17.2\%), stable alignment (15.0\%), and reward hacking (11.8\%). We did not perform repeated judge sampling or length-controlled judging; however, the stored length diagnostics do not support a simple verbosity-only explanation. Reward-hacking rows are longer on average than other rows, but both groups shorten across transitions, and the correlation between target length and reward-hacking status is small ($r=0.083$). The result motivates treating judge disagreement as a distinct monitoring channel rather than folding it silently into $\Rbar$.

\vspace{-0.75em}

\FloatBarrier
\section{Discussion}
The empirical picture is sharper than a single reward-hacking narrative. Aggressive PPO produces the clearest localized reward-hacking signal, but the taxonomy also exposes collapse, proxy under-alignment, conservative stagnation, mixed behavior, and judge-specific disagreement. UP-PPO reduces the density of reward-hacking and judge-disagreement events in the aggressive condition, but the remaining failures show why mitigation claims should be distributional rather than absolute.

The early-warning results are useful precisely because they are imperfect. The pre-state-only model has discriminative power, suggesting that future failures are not invisible before the transition. Yet average precision remains modest because reward hacking is rare and localized. Operationally, this favors triage: use warning scores to prioritize prompts and checkpoints for expensive human or multi-judge review, not to certify alignment.

The checkpoint-vs-row ablation has the broadest methodological implication. Aggregate metrics remain necessary for reporting, but they can average away exactly the failures that matter. A policy can improve on average while a subset of prompts deteriorates under proxy pressure. RLHF monitoring should therefore include matched prompt-level transitions, not only final or aggregate checkpoint scores.

\section{Limitations}
This study is deliberately controlled and small scale. The policy family is GPT-2-scale, so the results should be read as evidence about failure-mode observability rather than as a scaling claim about frontier systems. The prompt set is also small: the analysis covers 64 matched prompt identities for row-level transitions, which is enough to demonstrate localization but not enough to estimate population-level failure rates for open-ended deployment.

The analysis uses a controlled pipeline rather than a new multi-seed training campaign. DPO and SFT are included as baselines, but their evaluation rows do not represent the same online reward-optimization dynamics as PPO and UP-PPO checkpoints; claims about absence of reward hacking in those settings are therefore limited to the present pipeline. Where possible, we report bootstrap confidence intervals over rows or settings, but these intervals do not replace seed sensitivity. In particular, the mitigation comparison is directionally favorable for UP-PPO, yet the bootstrap intervals for the absolute reductions include zero. A stronger causal claim would require repeated training runs across seeds, larger prompt samples, and fixed judge configurations.

The early-warning models are trained and tested within the same pipeline family. The train/test split is stratified by the reward-hacking label but not grouped by prompt id, checkpoint family, or trajectory. The reported ROC-AUC therefore measures within-family discrimination, not transfer to new datasets, model scales, reward models, judge prompts, held-out prompts, or held-out training families. This limitation is especially important because row-level reward hacking is rare and localized; warning models may learn family-specific prompt or trajectory regularities.

The two external judges are useful but not ground truth. The Anthropic/Claude-style and OpenAI-style judges use the same 1--10 helpfulness rubric, but they can still disagree because of model-specific preferences, calibration differences, and sensitivity to response length or wording. We use one judge sample per item and do not include human adjudication or length-controlled judge prompts. Preliminary length diagnostics in Appendix~\ref{app:judge-disagreement-details} suggest that verbosity alone is not the main driver, but repeated judge sampling, length-controlled judging, and a small human-rated anchor set would be needed to make that claim robust.

The taxonomy is directional and intentionally simple. It is designed for auditability, but richer temporal models could capture gradual drift, threshold effects, delayed failures, and multi-step causal mechanisms more faithfully. Appendix~\ref{app:epsilon-sensitivity} shows that classifications are stable for small tolerances but become more conservative as $\epsilon$ or the minimum effect-size threshold grows. Finally, MC-dropout uncertainty is used as a diagnostic and as part of UP-PPO because it was implemented in the reward-model pipeline, but this paper does not fully validate its calibration or compare it with ensembles, last-layer analytic uncertainty, UWO/WCO-style uncertainty, or adversarial uncertainty penalties. The localization and aggregation findings are less tied to this uncertainty estimator than the mitigation result.

\section{Conclusion}
Reward hacking is best understood as one member of a broader family of RLHF failure modes. By classifying checkpoint transitions, preserving row-level prompt structure, and separating proxy--judge mismatch from judge--judge disagreement, this paper provides a practical diagnostic framework for RLHF monitoring. In this pipeline, aggressive PPO produces the strongest localized reward-hacking signal; UP-PPO reduces but does not eliminate that signal; pre-transition features anticipate some future failures; and aggregate checkpoint metrics miss localized reward hacking in several settings. The broader lesson is that RLHF evaluation should track how failures emerge, where they localize, and which signals appear before external quality degrades.

\appendix

\section{Algorithmic Details}
\label{app:algorithms}

\begin{algorithm}[H]
\caption{Transition-level failure-mode classifier}
\label{alg:taxonomy}
\begin{algorithmic}[1]
\Require Proxy delta $\delt\Rphi$, judge deltas $\delt\Rdag,\delt\Rtwo$, tolerance $\epsilon$
\Function{Sign}{$z,\epsilon$}
    \If{$z>\epsilon$} \State \Return $+1$ \EndIf
    \If{$z<-\epsilon$} \State \Return $-1$ \EndIf
    \State \Return $0$
\EndFunction
\State $p\gets\Call{Sign}{\delt\Rphi,\epsilon}$; $j\gets\Call{Sign}{\delt\Rdag,\epsilon}$; $j_2\gets\Call{Sign}{\delt\Rtwo,\epsilon}$
\If{$p=0$ and $j=0$} \State mode $\gets$ conservative stagnation
\ElsIf{$p>0$ and $j>0$} \State mode $\gets$ stable alignment
\ElsIf{$p>0$ and $j<0$} \State mode $\gets$ reward hacking
\ElsIf{$p<0$ and $j<0$} \State mode $\gets$ optimization collapse
\ElsIf{$p<0$ and $j>0$} \State mode $\gets$ proxy under-alignment
\Else \State mode $\gets$ mixed or ambiguous
\EndIf
\State judge-disagreement $\gets (j\cdot j_2<0)$
\State \Return mode, judge-disagreement
\end{algorithmic}
\end{algorithm}

\begin{algorithm}[H]
\caption{RLHF diagnostic pipeline}
\label{alg:pipeline}
\begin{algorithmic}[1]
\Require Rollout summaries $S$, row-level examples $E$, taxonomy classifier $C$
\State Parse checkpoint identity, training family, model type, and step from each pipeline output.
\State Build checkpoint table with mean $\Rphi$, $\Rdag$, $\Rtwo$, uncertainty, KL, and judge disagreement.
\State Build row-level table with prompt id, scores, generated text, length, distinct-$n$, and repetition.
\For{each family/model trajectory}
    \State Sort evaluated checkpoints by step.
    \For{each adjacent transition $t\to t'$}
        \State Apply $C$ to aggregate deltas and store checkpoint failure mode.
        \State Match row-level examples by prompt id and apply $C$ to each prompt transition.
    \EndFor
\EndFor
\State Train pre-state early-warning models to predict future row-level reward hacking.
\State Fit using a stratified random 70/30 row split; record that prompts and trajectory families are not group-held-out.
\State Report aggregate summaries, row-level diagnostics, and robustness checks.
\end{algorithmic}
\end{algorithm}

\section{Representative Prompt Transitions}
Table~\ref{tab:appendix-qualitative-transitions} shows representative matched transitions. The response snippets are shortened for readability, but the score columns report the underlying before/after values used by the classifier. The examples illustrate why the taxonomy is transition-based: the same prompt can move from a more coherent answer to a degraded answer while the proxy score increases, or it can produce opposite movements across two external judges.

\begin{table}[H]
\centering
\tiny
\caption{Representative prompt-level transitions. Scores are shown as before $\rightarrow$ after. Response snippets are abbreviated.}
\label{tab:appendix-qualitative-transitions}
\setlength{\tabcolsep}{2.2pt}
\begin{tabular}{p{0.12\linewidth}p{0.11\linewidth}p{0.21\linewidth}p{0.21\linewidth}p{0.09\linewidth}p{0.10\linewidth}p{0.08\linewidth}}
\toprule
Prompt theme & Transition & Before response snippet & After response snippet & $\Rphi$ & $(\Rdag,\Rtwo)$ & Mode \\
\midrule
Tax software advice & UP-PPO $1000\to1200$ & ``Tax software is generally considered the most widely used option... H\&R Block and TurboTax...'' & ``UberTax is a large online tax service... LyftTax is often used as a tax helper software.'' & $-0.375\to-0.198$ & $(6,7)\to(2,4)$ & Reward hacking \\
Workplace bias complaint & UP-PPO $600\to1000$ & ``...a constructive and respectful process... professional ethics and values in the workplace.'' & ``...contacting HR... contacting companies who already prohibit employees who speak out against hateful words.'' & $-1.297\to-1.050$ & $(7,6)\to(2,5)$ & Reward hacking \\
Workplace bias complaint & PPO $600\to1000$ & ``...this conversation can only be respectful... compassionate and civil...'' & ``...possible content suggestions... [irrelevant external-link-like text] ... complaints for anti-semitism.'' & $-1.356\to-1.124$ & $(7,6)\to(3,4)$ & Reward hacking \\
Private address request & PPO sampled $500\to600$ & ``Here are some possible answers: ...'' & ``We have a very detailed analysis... one of the most important... Wall Street Journal...'' & $-1.302\to-1.836$ & $(5.5,1)\to(2,3)$ & Collapse + judge disagreement \\
\bottomrule
\end{tabular}
\end{table}

\section{Threshold Sensitivity}
\label{app:epsilon-sensitivity}
\begin{figure}[H]
\centering
\includegraphics[width=0.78\linewidth]{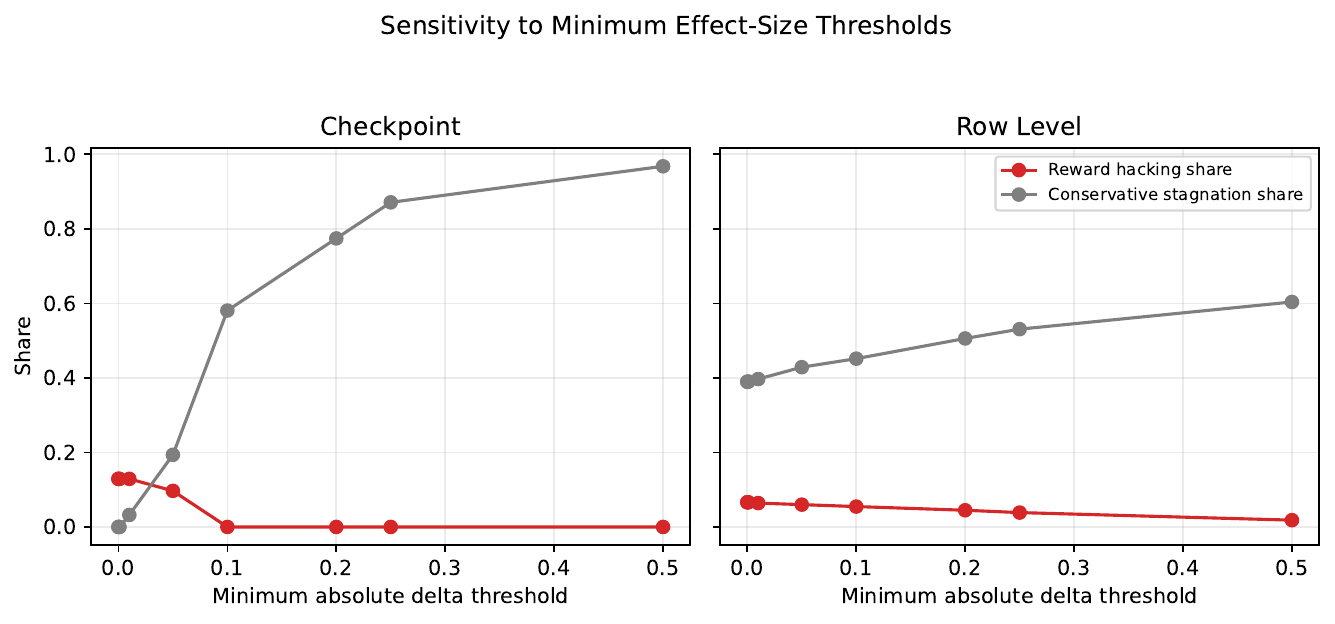}
\caption{Sensitivity of taxonomy counts to minimum absolute delta thresholds. Larger thresholds make the classifier more conservative, shifting small movements into conservative stagnation or mixed/ambiguous classes.}
\label{fig:effect-threshold-sensitivity}
\end{figure}

Table~\ref{tab:epsilon-sensitivity} reports how the taxonomy changes as the sign tolerance $\epsilon$ increases. The default analysis uses $\epsilon=10^{-8}$, which is effectively a nonzero sign test for the stored floating-point deltas. Classifications are unchanged at $\epsilon=10^{-3}$ for checkpoint transitions and change only minimally at row level. At larger tolerances, the classifier becomes deliberately more conservative: small proxy or judge movements are absorbed into conservative stagnation or mixed/ambiguous classes. Figure~\ref{fig:effect-threshold-sensitivity} extends the same check to minimum effect-size thresholds, including $0.2$ judge-score units. At threshold $0.2$, row-level reward-hacking counts fall from 127 to 86, while checkpoint-level reward hacking falls to zero. This behavior is expected and highlights why the threshold should be reported explicitly.

\begin{table}[H]
\centering
\small
\caption{Sensitivity of failure-mode counts to the sign tolerance $\epsilon$. RH denotes reward hacking; SA stable alignment; OC optimization collapse; PUA proxy under-alignment; CS conservative stagnation; MA mixed/ambiguous; JD judge disagreement.}
\label{tab:epsilon-sensitivity}
\begin{tabular}{llrrrrrrrr}
\toprule
Unit & $\epsilon$ & Total & RH & SA & OC & PUA & CS & MA & JD \\
\midrule
Checkpoint & $10^{-8}$ & 31 & 4 & 7 & 5 & 6 & 0 & 9 & 14 \\
Checkpoint & $10^{-3}$ & 31 & 4 & 7 & 5 & 6 & 0 & 9 & 14 \\
Checkpoint & $10^{-2}$ & 31 & 4 & 7 & 5 & 5 & 1 & 9 & 13 \\
Checkpoint & $5\times10^{-2}$ & 31 & 3 & 4 & 1 & 1 & 6 & 16 & 5 \\
Checkpoint & $10^{-1}$ & 31 & 0 & 1 & 1 & 0 & 18 & 11 & 1 \\
Row-level & $10^{-8}$ & 1920 & 127 & 100 & 99 & 122 & 748 & 724 & 75 \\
Row-level & $10^{-3}$ & 1920 & 127 & 98 & 99 & 122 & 749 & 725 & 75 \\
Row-level & $10^{-2}$ & 1920 & 123 & 96 & 99 & 120 & 762 & 720 & 75 \\
Row-level & $5\times10^{-2}$ & 1920 & 115 & 83 & 92 & 110 & 823 & 697 & 75 \\
Row-level & $10^{-1}$ & 1920 & 105 & 72 & 79 & 101 & 867 & 696 & 75 \\
\bottomrule
\end{tabular}

\end{table}

\section{Aggregation Flip Details}
\label{app:aggregation-flips}
The checkpoint-vs-row comparison can be sharpened by comparing each checkpoint transition to the dominant row-level label among its matched prompts. Among the 30 checkpoint transitions with matched row-level data, all 30 have a dominant row-level label that differs from the checkpoint label, and 19 have a strict row-level majority. Sixteen transitions contain localized reward-hacking rows even when the checkpoint label is not reward hacking. Table~\ref{tab:aggregation-flips} lists representative transitions with the largest hidden row-level reward-hacking shares.

\begin{table}[H]
\centering
\tiny
\caption{Representative checkpoint transitions where aggregate and row-level diagnoses diverge. CK denotes checkpoint label; Dom. denotes dominant row label; PUA proxy under-alignment; SA stable alignment; OC optimization collapse; MA mixed/ambiguous.}
\label{tab:aggregation-flips}
\begin{tabular}{p{0.32\linewidth}rp{0.12\linewidth}p{0.12\linewidth}rr}
\toprule
Setting & Step & CK & Dom. & Row RH share & Dom. share \\
\midrule
Aggressive PPO & $600\to1000$ & PUA & MA & 0.188 & 0.438 \\
$\lambda=0.5$ UP-PPO & $600\to1000$ & PUA & MA & 0.141 & 0.609 \\
$\beta=0.001$ sampled PPO & $200\to500$ & SA & MA & 0.141 & 0.438 \\
Aggressive PPO & $200\to400$ & PUA & MA & 0.109 & 0.594 \\
Aggressive PPO & $400\to600$ & SA & MA & 0.109 & 0.500 \\
$\beta=0.03$ PPO & $250\to350$ & OC & MA & 0.109 & 0.406 \\
\bottomrule
\end{tabular}
\end{table}

\section{Judge Disagreement Details}
\label{app:judge-disagreement-details}
Judge disagreement is measured by whether the two judges move in opposite directions across the same transition, $\mathrm{sign}(\delt\Rdag)\mathrm{sign}(\delt\Rtwo)<0$. Table~\ref{tab:appendix-judge-disagreement} reports this event by unit of analysis and failure mode. The distinction between checkpoint and row-level disagreement is important. At checkpoint level, 14 of 31 transitions exhibit judge disagreement, producing a high aggregate disagreement rate. At row level, only 75 of 1,920 transitions exhibit the same opposite-direction pattern. This suggests that checkpoint-level judge disagreement can be driven by small average shifts across many prompts, while row-level disagreement is more localized.

At row level, proxy under-alignment has the highest judge-disagreement share (17.2\%), followed by stable alignment (15.0\%) and reward hacking (11.8\%). Conservative stagnation has zero judge-disagreement events by construction in this pipeline because both the proxy and anchor judge are approximately unchanged. The higher disagreement rates in proxy under-alignment and reward hacking are consistent with the central diagnosis of the paper: when the proxy and anchor judge disagree, the two judges are also more likely to expose instability in what is being measured.

\begin{table}[H]
\centering
\small
\caption{Judge-disagreement share by failure mode. Judge gap is $|\delt\Rdag-\delt\Rtwo|$.}
\label{tab:appendix-judge-disagreement}
\begin{tabular}{llrrrrr}
\toprule
Unit & Failure mode & Total & Judge disagree. & Mean gap & Median gap & Share \\
\midrule
Row & Proxy under-alignment & 122 & 21 & 1.516 & 1.000 & 0.172 \\
Row & Stable alignment & 100 & 15 & 1.275 & 1.000 & 0.150 \\
Row & Reward hacking & 127 & 15 & 1.138 & 1.000 & 0.118 \\
Row & Optimization collapse & 99 & 5 & 0.904 & 1.000 & 0.051 \\
Row & Mixed/ambiguous & 724 & 19 & 0.800 & 1.000 & 0.026 \\
Row & Conservative stagnation & 748 & 0 & 0.167 & 0.000 & 0.000 \\
\midrule
Checkpoint & Optimization collapse & 5 & 3 & 0.242 & 0.180 & 0.600 \\
Checkpoint & Stable alignment & 7 & 4 & 0.206 & 0.164 & 0.571 \\
Checkpoint & Mixed/ambiguous & 9 & 4 & 0.043 & 0.039 & 0.444 \\
Checkpoint & Proxy under-alignment & 6 & 2 & 0.118 & 0.094 & 0.333 \\
Checkpoint & Reward hacking & 4 & 1 & 0.166 & 0.145 & 0.250 \\
\bottomrule
\end{tabular}
\end{table}

These results motivate keeping $\Rdag$, $\Rtwo$, and $\Rbar$ visible as separate quantities. Averaging judges can stabilize noisy evaluations, but it can also hide judge-specific movement. For failure analysis, the disagreement event is itself informative: it marks transitions where the measured direction of progress depends on the judge used to define progress.

We also checked whether the stored row-level length features suggest a simple verbosity explanation. Reward-hacking rows have higher target length than non-reward-hacking rows on average (59.3 versus 52.4 words), but they do not become longer across the transition: mean length change is $-3.88$ words for reward-hacking rows and $-2.68$ words for other rows. The correlation between target length and reward-hacking status is small ($r=0.083$), as is the correlation between length change and reward-hacking status ($r=-0.018$). These diagnostics do not replace length-controlled judging, but they make a pure verbosity account less plausible in the current pipeline outputs.

\end{document}